\documentclass[preprint,nopreprintline]{elsarticle}
\usepackage{amsmath,amssymb}

\DeclareMathOperator{\nh}{N}
\DeclareMathOperator{\front}{NF}

\usepackage{xcolor}
\usepackage{rotating,multirow}

\begin{document}

\begin{frontmatter}

\title{Evaluating the Quality of Finite Element~Meshes with Machine Learning}

\author{Joachim Sprave}

\author{Christian Drescher}

\address{Mercedes-Benz AG}

\begin{abstract}
This paper addresses the problem of evaluating the quality of finite element meshes for the purpose of structural mechanic simulations. It proposes the application of a machine learning model trained on data collected from expert evaluations. The task is characterised as a classification problem, where quality of each individual element in a mesh is determined by its own properties and adjacency structures. A domain-specific, yet simple representation is proposed such that off-the-shelf machine learning methods can be applied. Experimental data from industry practice demonstrates promising results.
\end{abstract}

\end{frontmatter}

\section{Introduction}
Finite element methods (FEM) play an important role in the digital product development process of many industries, including aerospace and automotive. FEM allow to complement or replace physical experiments in the development process by numerical simulations on a discretization of the design's smooth geometry into a \emph{mesh} of a finite number of polygons called \emph{elements} that connect points sampled from the geometry, called \emph{nodes}.

Most commercially successful mesh generation is based on variants of the decompositional \emph{medial axis algorithm} or the recursive \emph{advancing front algorithm} that have been tailored to industrial applications based on years of experience. There is a significant body of literature on mesh generation. For a general overview, we refer to \cite{SO2000}, and to \cite{Bommes:2013:QGP:2771539.2771544} for a recent literature survey on mesh generation with quadrilateral elements. Notable approaches that work globally rather than decompositional or recursive are~\cite{KNP07} and~\cite{PS97}.

An important aspect to the practical application of FEM is the quality of meshes with respect to their fitness to enable accurate simulations. A lot of effort is put into formalising quality criteria to guide mesh generation algorithms towards high-quality meshes, albeit with mixed success in the presence of complex and sometimes conflicting requirements from different simulation disciplines. (The advantage of being able to work on a mesh that is shared among different disciplines justifies the effort required to generate joint mesh.)
%
Hence, it is left to experienced engineers to review meshes and rework individual elements or adjacent sets of elements (i.e., \emph{subareas}) of meshes in manual, time-consuming work. The cognitive process underlying this highly subjective activity has yet to be formalised into a set of requirements or heuristics for mesh generation.

We propose an alternative, data-driven approach to evaluating the quality of meshes by means of training a machine learning model on data from expert evaluations that is able to generalise to unseen data. An advantage of this approach is that machine learning algorithms can abstract from the low-level features represented as input data to higher-level concepts, e.g., represented by hidden layers in an artificial neural network~\cite{ruhiwi86}. A machine learning model with good fit and generalisation can be applied to review and rework unseen, future meshes. Where quality is predicted unsatisfactory by the model, undesirable mesh structures may be generated anew and re-evaluated until a fixed point is reached through a fully automated routine.

However, the application of machine learning to mesh evaluation is non-trivial. For one, the explicit graph structure of meshes makes their evaluation different from traditional machine learning tasks. For instance, the quality of a subarea of a mesh is often interdependent with structure and quality of the remaining mesh, whilst traditional machine learning tasks consider independent observations. For another, meshes are generally unstructured, i.e., the number of elements sharing a node is not constant. This rules out direct constant-size vector-like representations that are required for most off-the-shelf machine learning algorithms. There are many different avenues to represent the task of mesh evaluation as a machine learning problem, including projection to 2D images or neighbourhood aggregation. This paper follows the latter approach.
%
The contributions of this paper are many-fold.

First, we characterise the problem of evaluating mesh quality as a classification problem for machine learning. For this purpose, we introduce the element neighbourhood graph induced by a mesh, where each element is represented by a vertex, and edges between vertices represent adjacent elements. Then, the machine learning task is to predict whether an element belongs to a subarea that requires rework in order to achieve acceptable mesh quality.

Second, as most off-the-shelf machine learning methods work on vector-like representations rather than graph structures, we propose to extract simple but domain-specific, statistical features from the neighbourhood of each element. We will rely on machine learning to abstract from our low-level features to higher-level concepts.

Third, we conducted an empirical study of our approach that includes meshes from parts of a real-world passenger vehicle. Experimental results demonstrate applicability but also some limitations of our approach.

The remainder of this paper is organised as follows: Section~\ref{sec:prelim} introduces necessary background. Section~\ref{sec:eng} introduces how we capture the neighbourhood of an element on a mesh, and in Section~\ref{sec:encoding} we show how we represent the task of evaluating a mesh for supervised machine learning. We demonstrate practical applicability in Section~\ref{sec:experiments}. Then, we outline and discuss some alternative machine learning approaches to mesh evaluation in Section~\ref{sec:discussion}. Finally, the paper is concluded in Section~\ref{sec:conclusions}.

\section{Preliminaries} \label{sec:prelim}
Machine learning studies computer algorithms that optimise the parameters of a (mathematical) \emph{model} to fit data with respect to some cost function. This process is also called \emph{learning} or \emph{training} a machine learning model. Supervised (machine) learning is the task of optimizing the parameters~$\theta$ of a function~$f_\theta(x)$ that maps \emph{input} data points described by the values to the attributes of a \emph{feature vector}~$x$ to an output, called \emph{label}, based on sample pairs of input and label. If the model is able to generalise to unseen pairs, i.e., succeeds predicting the true label of validation data with little error, it may be applied to make predictions about unseen inputs. When labels represent group membership, the task of identifying the label for any given input is called a \emph{classification problem}. Many types of models that are suitable for supervised learning on classification problems are known to the literature, including classification trees and feedforward neural networks.

A classification tree~\cite{brfrolst84} is a directed tree-like model that can be constructed from data samples by recursively selecting an attribute from the feature vector and partitioning the samples into subsets according to the (discretised) values taken for the attribute. Hence, the parameters~$\theta$ of a classification tree model include attribute selections and discretisation of attribute values for each partitioning. Extremely randomised trees~\cite{geerweh06}, for instance, randomise strongly both attribute selection and attribute discretisation. A classification tree model is applied to unseen input by recursively following the partitioning (\emph{branches}) that matches to the values for the attributes of the input. The (predicted) label of the input data is the majority label of the partition where recursion aborts. In order to control predictive performance, the folklore of machine learning practice also considers thresholds on the probability estimate of class membership as an alternative to majority label.
%
Better generalisation to unseen data may be achieved by combining multiple models~\cite{zhou2012}.

A feedforward (artificial) neural network is a directed acyclical graph-like model that consists of multiple layers of (computational) units, where each unit from one layer has connections to only units of the subsequent layer. Connections represent a flow of information, where the output of a unit is input to another unit whenever the former is connected to the latter. There is precisely one unit per attribute of the feature vector that make up the first layer in the network, one unit per attribute. Their output is provided by the respective values of any input data. The output of each unit from all other layers is computed by a linear combination of their input values that is passed through a non-linear activation function. The output of the units from the last layer represent the label associated with the input data. The parameters~$\theta$ of a feedforward neural network include all coefficients to the linear combination computed by each unit. 
%
A feedforward neural network can be trained, e.g, by first defining its topology (or architecture), initialising~$\theta$ with (random) values, and then adjusting~$\theta$ based on sample input in a process called backpropagation~\cite{ruhiwi86}.

An important aspect of machine learning is the design of the feature vector. Vector- or grid-like data, where each data point (e.g., row in a data table) takes values for a known fixed set of attributes, are referred to as structured data. Structured data can be easily exploited for machine learning because a feature vector can be directly modelled after the available data columns.
A much bigger challenge is finding a way to encode unstructured data, like graphs, into a feature vector for machine learning. A common limitation is that the structure of any feature vector is defined a-priori and, in particular, cannot include data of arbitrary size. However, the number of vertices that share an edge with any given vertex in a graph can be arbitrary in general.

This paper applies the usual basic graph theory definitions and notations, where a finite and undirected graph~$\mathcal{G} = (V, E)$ with no loops or multiple edges is defined by a finite set of vertices~$V$ and a set of edges~$E$, each of which is a set of two vertices from~$V$. Instead of~$v \in V$ and $\{v, w\} \in E$ we also use the notation~$v \in \mathcal{G}$ and $(v, w) \in \mathcal{G}$. For any vertex~$v \in \mathcal{G}$, the vertex degree of~$v$ is $|\{w \mid  (v,w) \in \mathcal{G}\}|$, i.e., the number of vertices that share an edge with~$v$. A graph~$\mathcal{G}$ is called \emph{regular} if the vertex degree of every vertex in~$\mathcal{G}$ is the same, and \emph{irregular} otherwise.

Meshes encode unstructured data closely related to graphs.
%
In the context of structural mechanic simulations, meshes consist of elements that connect only few nodes (for performance reasons). Whilst the smallest possible shape for this purpose is a triangle, however, quadrilaterals are preferred over triangles for numerical reasons that go beyond the scope of this paper.
%
Hence, in the following, we consider \emph{quad-dominant} meshes, i.e., meshes that consist of mostly quadrilateral elements and few triangles. Given a finite set of points $\mathcal{N} \subset \mathbb{R}^3$ sampled from some geometry, an \emph{element} is either a triangle from~$\mathcal{N}^3$ or a quadrilateral from~$\mathcal{N}^4$. Then, a \emph{(quad-dominant) mesh} is a finite set $\mathcal{M} \subseteq \mathcal{N}^3 \cup \mathcal{N}^4$.

The choice of mesh to approximate the geometry of a design directly affects accuracy or stability of subsequent FEM. In some of our application scenarios, the ideal mesh consists of coplanar quadratic elements. While it is unrealistic to approximate most relevant geometries with such ideal, triangles are introduced where quadrilaterals would be distorting. 
%

Various metrics to measure \emph{element quality} and \emph{mesh quality} are employed to guide mesh generation towards favorable results. Aspect ratio, skewness, and warpage are examples of element quality metrics from mesh generation folklore. The \emph{aspect ratio} of an element is the aspect ratio of a minimum rectangle containing the element. \emph{Skewness} is the angular difference of the medians of the opposing edges in a quadrilateral. \emph{Warpage} is a measure that quantifies how much the nodes of a quadrilateral divert from being coplanar. Examples for mesh quality metrics are the minimum edge length of any element, or the fraction of triangles in a quad-dominant mesh.

\begin{figure}[ht]
    \centering
    \includegraphics[width=3.0in]{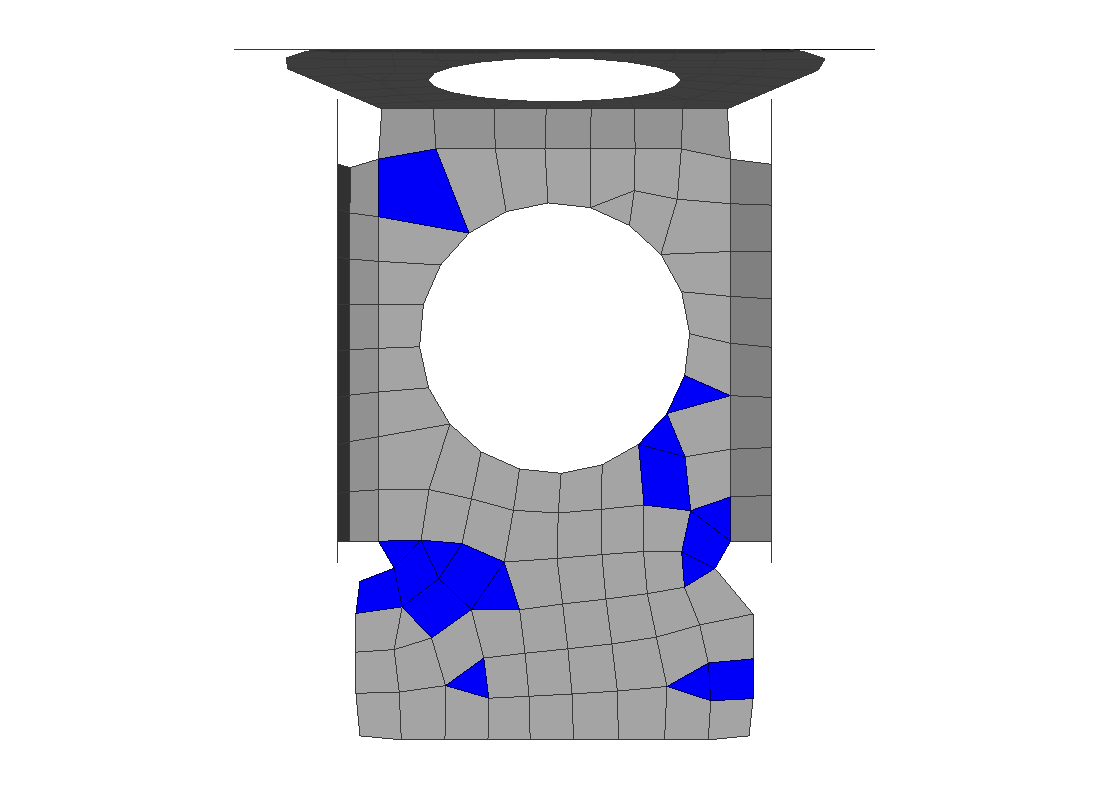}
    \caption{Illustration of the results from an evaluation of mesh quality of a quad-dominant mesh for a sheet metal part. Blue shaded elements are labelled for rework because they are considered inconsistent with (informal) requirements for structural mechanic simulations.}
    \label{fig:simplelabeled}
\end{figure}

Whilst mesh generation optimises such quality metrics, industry practice knows many additional, often conflicting requirements on mesh quality from different FEM disciplines. Hence, every relevant mesh undergoes a structured review process, where experienced engineers identify and label sets of elements that require rework in order to achieve a mesh that enables accurate structural mechanic simulations. An example is provided in Figure~\ref{fig:simplelabeled}. Rework considers the modification of elements, including the generation of an alternative mesh, until a result is achieved that is deemed satisfactory by the reviewer.

\section{Element Neighbourhood Graph} \label{sec:eng}
\begin{figure}
    \centering
    \includegraphics[width=4.0in]{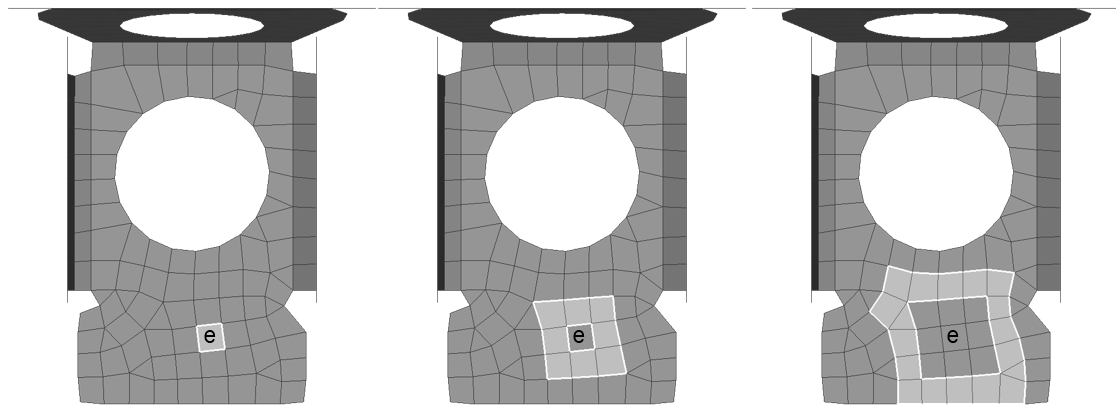}
    \caption{Illustration of an element marked by~$e$ and its $k$-ring neighbourhood frontier in bright shade, for $k = 0,1,2$ (from left to right).}
    \label{fig:nbh}
\end{figure}

When classifying an element or a set of elements in any given mesh, expert reviewers naturally consider their fit within the remaining mesh, and nearby elements in particular.
In order to capture adjacency of elements, we introduce the \emph{(element) neighbourhood graph} induced by a mesh, where each element is identified by a vertex, and an edge connects two vertices whenever the respective elements share a node. Formally, the element neighbourhood graph of a mesh~$\mathcal{M}$ is the (undirected) graph defined by
\[
    \mathbf{NG}(\mathcal{M}) = (\mathcal{M}, \{ (\varepsilon, \varepsilon^\prime) \mid \varepsilon,\varepsilon^\prime\in\mathcal{M},\ \varepsilon\neq\varepsilon^\prime,\    \varepsilon\cap\varepsilon^\prime\neq\emptyset \})\text{.}
\]
%
%
%
For any element~$\varepsilon$, the set of elements that consists of~$\varepsilon$ and all elements that share a node with~$\varepsilon$ is called the (1-ring) neighbourhood of~$\varepsilon$. The neighbourhood of~$\varepsilon$ can be expanded to its 2-ring neighbourhood by recursively including the neighbourhood of each member.
Formally, for any graph~$\mathcal{G}$ the \emph{k-ring neighbourhood} of a vertex~$v \in \mathcal{G}$ for any integer~$k \in \mathbb{N}_0$ is defined by
\[
\nh^k_{\mathcal{G}}(v) =
\begin{cases}
\{v\} & \text{for } k = 0\text{,} \\
\nh^{k-1}_{\mathcal{G}}(v) \cup \{w \mid u \in \nh^{k-1}_{\mathcal{G}}(v) \land (u,w) \in \mathcal{G}\} & \text{for } k > 0\text{.}
\end{cases}
\]
In order to access vertices in the $k$-ring neighbourhood of~$v$ that are not included in its $k-1$-ring neighbourhood, we define the \emph{$k$-ring neighbourhood frontier} of~$v$ for any integer~$k \in \mathbb{N}_0$ by
\[
\front^k_{\mathcal{G}}(v) =
\begin{cases}
\nh^0_{\mathcal{G}}(v) & \text{for } k = 0\text{,} \\
\nh^k_{\mathcal{G}}(v) \setminus \nh^{k-1}_{\mathcal{G}}(v) & \text{for } k > 0\text{.}
\end{cases}
\]
Accordingly, the \emph{k-ring (element) neighbourhood} of an element~$\varepsilon\in\mathcal{M}$ is given by~$\nh^k_{\mathbf{NG}(\mathcal{M})}(\varepsilon)$, and the the \emph{k-ring (element) neighbourhood frontier} of an element~$\varepsilon\in\mathcal{M}$ is~$\front^k_{\mathbf{NG}(\mathcal{M})}(\varepsilon)$. 
Examples are provided in Figure~\ref{fig:nbh}.
%

\section{Element Classification} \label{sec:encoding}
In order to automate the task of reviewing a mesh for necessary rework, we formulate a classification problem for supervised learning. 
Our formulation exploits the reasonable assumption that the (binary) label of each individual element (i.e., rework or not) in a mesh can be determined by selected properties and adjacency structures. Whilst the precise relationship remains unclear, we apply machine learning to fit a model on historical data from expert evaluations.

To begin with, we associate the label of an element with properties of the element that can be used to guide its classification, i.e., we define the attributes of a feature vector. An obvious choice are properties of the element itself, and includes information that is known for most elements, such as aspect ratio and skewness. The label of an element might also be dependent on features from other elements in the mesh. (This contrasts element classification to traditional machine learning tasks, where every observations is considered independent from others.) In order to capture the neighbourhood of an element, we want to include additional features based on adjacency as captured by the neighbourhood graph. Since a node in a mesh can be shared by an arbitrary number of elements, the neighbourhood graph of a mesh is irregular in general. This provides a challenge for the direct application of many machine learning techniques which naturally handle feature vectors of a fixed size.

A straightforward approach to represent irregular graphs for the application of machine learning is to collect information from neighbouring vertices and aggregate the values of each relevant feature. Following this idea, we characterise the neighbourhood of an element by collecting and aggregating the values of quality metrics from the elements of its $k$-ring neighbourhood frontier at varying distance~$k \leq \mathbf{k}$ for some limit~$\mathbf{k} \in \mathbb{N}_0$.

Let $\pi_i: \mathcal{M} \to \mathbb{R}$ for $i \in 1, \dots, m$ be a family of real-valued properties associated with each element in~$\mathcal{M}$, e.g., element quality metrics like warping or aspect-ratio, and $\phi_j:  \mathcal{P}(\mathbb{R}) \to \mathbb{R}$ for $j \in 1, \dots, n$ be a family of aggregation functions, e.g., minimum or mean. For any~$\mathbf{k} \in \mathbb{N}_0$, we define the (constant-size) \emph{feature tensor associated with an element}~$\varepsilon\in\mathcal{M}$ by
\[
((\phi_j(\{\pi_i(\varepsilon^\prime) \mid \varepsilon^\prime \in \front^k_{\mathbf{NG}(\mathcal{M})}(\varepsilon)\}))_{1 \leq i \leq m, 1 \leq j \leq n})_{0 \leq k \leq \mathbf{k}}\text{.}
\]
The concrete choice of properties~$\pi_i$, aggregation methods~$\phi_j$, and limit~$\mathbf{k}$ represent hyperparameters to modelling the classification problem. Having three axis, the feature tensor can be reshaped to a feature vector of size~$(\mathbf{k}+1)mn$.

Note that our feature vector includes only simple, low-level properties of elements and, in particular, does not cover properties defined between adjacent elements (e.g., the dihedral angle) or more complex properties defined on sets of neighbouring elements. In our approach, however, we rely on machine learning to abstract from the low-level features to higher-level concepts, e.g., represented by hidden layers in a feedforward neural network or branches in classification trees.

\section{Experimental Results} \label{sec:experiments}
In this section, we report the results of extensive experiments to demonstrate the applicability and effectiveness of our approach on industry-standard meshes generated from a real-world design.
The objects considered in our analysis include 317 parts of the body-in-white (vehicle body without engine, chassis subframe, electronics, interior fittings, etc.) of a recent Mercedes-Benz passenger car.
We have limited our study to \emph{shell} meshes, sometimes called surface meshes. Shell meshes are of particular importance to industry practice because they are used to simulate the behavior of sheet metal, plastic, and composite parts based on the theory of thin shells. We refer to \cite{Olszak1980} for further reading on this topic.
Each mesh was generated by a commercial application to industry-standard and without human interaction, and evaluated by a domain expert. In turn, the expert marked sets of elements for rework whenever their quality (e.g., fitness to enable accurate simulations) was considered unsatisfactory.

The expert evaluation serves as a class label for each element, i.e., marked elements are labelled~\emph{rework}, and the remaining elements are labelled as~\emph{passed}. Note that there is an obvious caveat: The semantics of an element being marked by an expert is not necessarily that of a lack of individual (element) quality, but rather that of belonging to a set of elements that, in combination, might require rework. Adding to the label noise is that, sometimes, domain experts have marked a small set of adjacent elements even though a single element was intended, and very often, rows of otherwise inconspicuous elements were marked to connect low quality elements.

Yet, only a small fraction of elements have been considered for rework by domain experts, as industry-standard mesh generation has been tailored with years of experience. In fact, we observe an extreme class imbalance: only $45819$ of the $1667719$ elements in our study ($2.75\%$) have been labelled for rework. The ``worst'' mesh consists of $64$ elements of which $94 \%$ are labelled for rework, whilst two meshes contain only elements labelled as passed. 

We have extracted seven low-level properties from elements. Five of which are continuous attributes, namely skewness, aspect ratio, warping, area, and the angle between the surface normals (through the centre of mass of an element and other elements from its neighbourhood) as a measure of curvature. The two remaining attributes are boolean, indicating triangle and border elements (elements that have at least one edge that is not shared with another element). We have fixed aggregate statistics to minimum, maximum, and mean, and the limit~$\mathbf{k} = 4$ as the other hyperparameters for the construction of feature tensors. Initial tests showed that increasing~$\mathbf{k}$ did not improve predictive performance, presumably because little useful information can be aggregated from $k$-ring neighbourhood frontiers with increasing~$k$ as such sets generally include an increasing amount of elements at increasing distance and from opposing directions. Altogether, the feature tensor consists of 84 attributes.

Our experiments considered extremely randomised trees~(ExtraTrees) and a feedforward neural network~(FNN). The ExtraTrees setting considers an ensemble of $100$~extremely randomised trees, each with access to only $\sqrt{84}$~randomly selected attributes for partitioning. The predicted label to an input was determined by averaging the probability estimate of the trees for the input belonging to the rework class, and applying a threshold.
%
The FNN setting considers a feedforward neural network with three hidden layers of sizes $64/128/16$, respectively, ReLU activation on all units in the hidden layers, and sigmoid activation on the single unit in the output layer. Batch normalisation was applied after the first and second hidden layer. This architecture was selected via limited experimentation. Network optimisation was performed using the Adam optimiser with the learning rate set to~$0.01$.

In order to assess how accurately our approach performs as a predictive model in practice, i.e., how well it generalises to unseen meshes, we have implemented our experiments using $10$-fold crossvalidation, a commonly applied technique to evaluate machine learning models on limited sample size.
To begin with, the 317 meshes were randomly partitioned into $10$~roughly equal sized subsets. Then, the label of each element from the meshes included in each subset were predicted by a machine learning model trained only on data from the remaining subsets. Note that we have decided to partition element data by mesh in order to avoid information leakage since the label of elements in the same mesh are generally not independent. As a result, the number of elements in each crossvalidation subset can vary strongly with the size of the meshes included in the subset. In fact, the smallest mesh considered in our experiments consists of $21$ elements and the largest mesh consists of $108759$ elements. The median mesh consists of $1984$ elements. Therefore, we report predictive performance by collecting individual predictions from all crossvalidation subsets (as if predictions stem from a single experiment) instead of averaging performance metrics across crossvalidation subsets.

A summary of our experimental results is given in Table~\ref{tab:results}. In order to control predictive performance and to demonstrate the available tradeoff between different metrics, we have included varying thresholds for the output of ExtraTrees and FNN, respectively, into our experiments.
\begin{table}
    \centering
\begin{tabular}{|c|c|c|c|c|c|c|c|c|}

\multicolumn{9}{c}{ExtraTrees}\\
\hline
th&TP\%&TN\%&FP\%&FN\%&Precision&Recall&Acc.&F1-score\\
\hline
0.25&2.84&86.25&10.75&0.16&0.21&0.95&0.89&0.34\\
0.50&2.32&92.45&4.55&0.68&0.34&0.77&0.95&0.47\\
0.75&1.32&95.78&1.22&1.68&0.52&0.44&0.97&0.48\\
\hline
\multicolumn{9}{c}{}\\
\multicolumn{9}{c}{FNN}\\
\hline
th&TP\%&TN\%&FP\%&FN\%&Precision&Recall&Acc.&F1-score\\
\hline
0.25&2.63&88.76&8.25&0.37&0.24&0.88&0.91&0.38\\
0.50&2.32&91.53&5.47&0.68&0.30&0.77&0.94&0.43\\
0.75&1.77&93.98&3.03&1.23&0.37&0.59&0.96&0.45\\
\hline
\end{tabular}
    \caption{Performance comparison results with respect to shares of true positives (TP\%; correctly predicted for {rework}), true negatives (TN\%; correctly predicted as {passed}), false positives (FP\%; misclassified for {rework}), false negatives (FN\%; misclassified as {passed}), respectively, and precision, recall, accuracy, and F1-score for the ExtraTrees and FNN settings at varying thresholds (th).}
    \label{tab:results}
\end{table}
\begin{figure}
    \centering
    \includegraphics[width=4.0in]{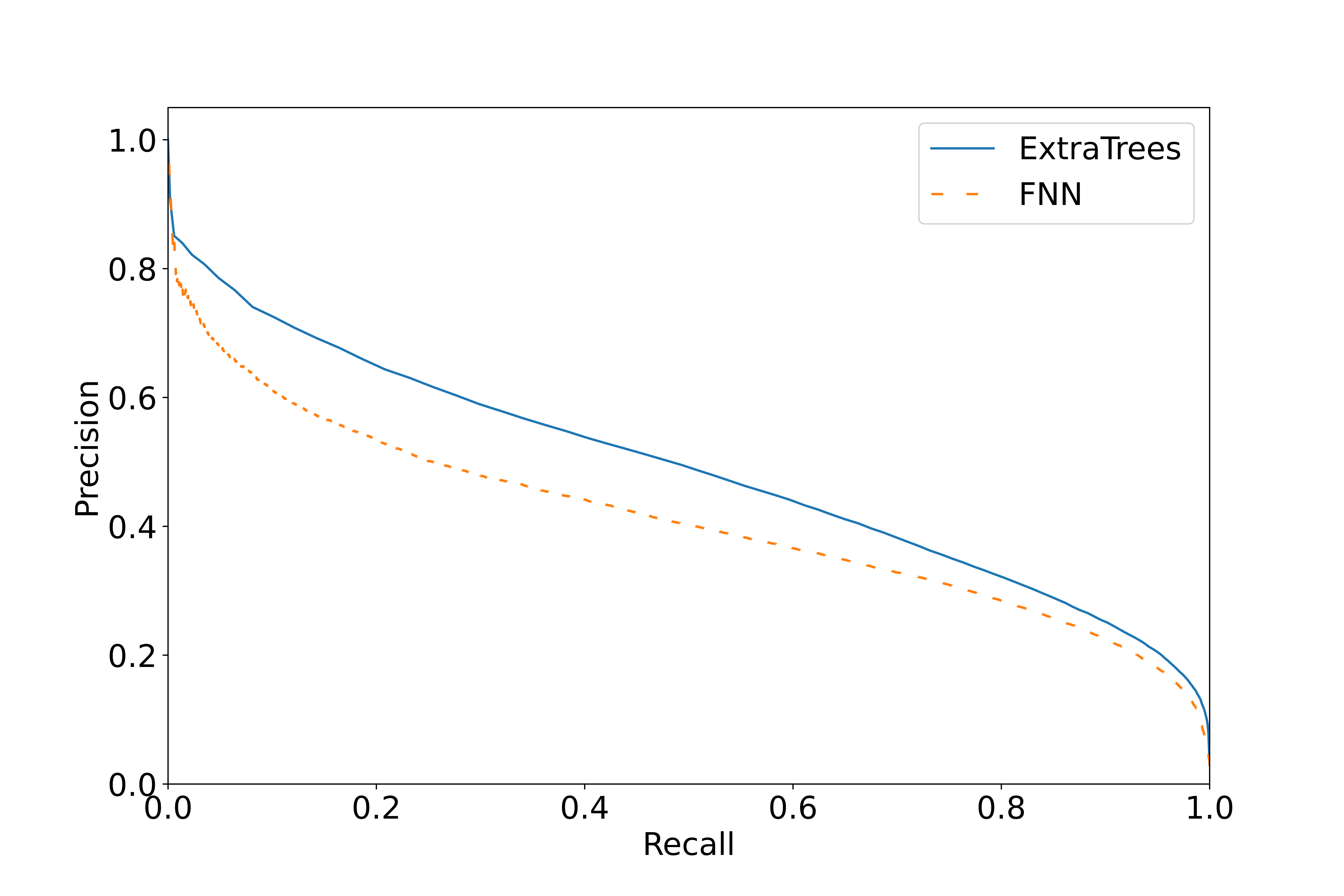}
    \caption{Precision/recall curve for ExtraTrees and FNN illustrates that the predictive performance of the ExtraTrees setting is superior to the FNN setting.}
    \label{fig:curve}
\end{figure}

Given the stark class imbalance in our experimental data, accuracy can be a misleading metric. Any trivial model for our application, e.g., one that predicts \emph{passed} for every input, can achieve an accuracy of $97.25\%$. In fact, both ExtraTrees and FNN perform only slightly below this value.
Hence, we turn our attention to the precision and recall statistics. Precision and recall are of particular practical relevance to our application domain, where recall measures the probability of detecting an element that expert review marks for rework, and precision measures the probability an element that is predicted to require rework is also marked by expert review. (Note that the aforementioned trivial model achieves zero recall.) With the threshold set to~$0.50$, ExtraTrees and FNN correctly identify approximately $77\%$ of the elements that were marked for rework. Whilst demonstrating some success, these are unfavourable results: both models misclassify nearly $23\%$ of the elements that require rework, yielding meshes that might hinder accurate simulations. Lowering the threshold hyperparameter to~$0.25$, however, lifts recall to approximately $95\%$ and $88\%$, respectively. This comes at the price of reduced precision, from approximately $34\%$ down to $21\%$ and $30\%$ down to $24\%$, respectively, indicating a rise in false positives. A more detailled view on the available tradeoff between precision and recall is provided by the diagram in Figure~\ref{fig:curve}. Depending on what cost is associated with false positives, our results demonstrate practical applicability.

\begin{figure}
    \centering
    \includegraphics[width=4.0in]{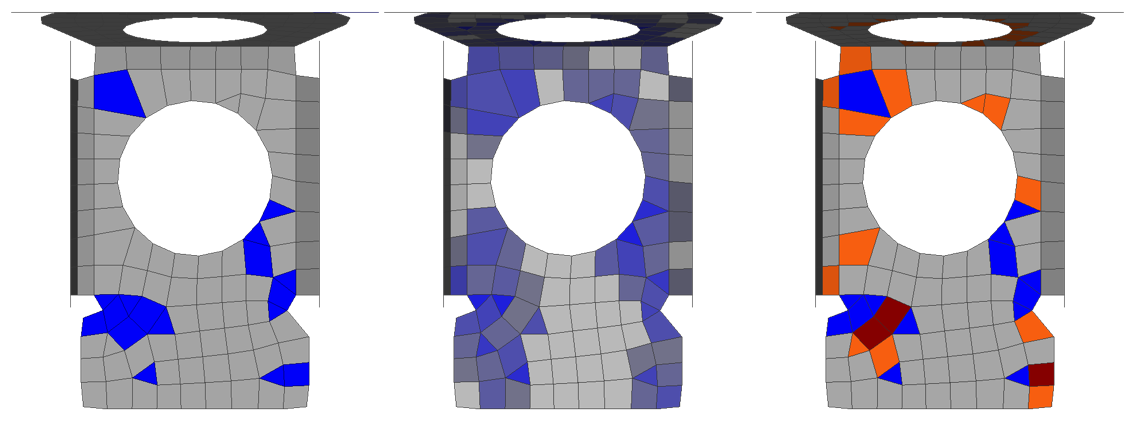}
    \caption{Illustration of ground truth labels (left; blue shade marks elements labelled for rework), probability estimate of being labelled for rework from applying ExtraTrees (center; shades of blue), and a detailled evaluation of the predictive performance with respect to the identification of elements that require rework when fixing the threshold at~$0.4$ (right; blue = true positives, grey = true negatives, orange = false positives, red = false negatives) for a sample mesh from our experiments. Allowing for slight deviations from the ground truth label, the ExtraTrees setting at this threshold correctly identifies some elements from each set of adjacent elements that are labelled for rework, and only predicts rework for one subarea of the mesh that is was not labelled as such in the expert review.}
    \label{fig:example}
\end{figure}

We believe that, by large parts, measurable performance is limited by the vague semantics of an element being marked for rework by expert review, as per our previous reservations about label noise. Visual inspection of the predicted labels reveals, however, that both models roughly identify many subareas of a mesh that require rework to a degree that warrants practical applicability. An example is provided in Figure~\ref{fig:example}. We conclude that using an element-wise evaluation of predictive performance provides structurally unfavourable results, and their conclusiveness with respect to successful completion of the original objective to identify subareas that require rework is limited.
%
%

Hence, in a separate experiment, we have presented the results of our approach to two groups of engineers: experienced engineers who have participated in the review process for our previous experiments, and novice engineers from a support team that is not specialised on this particular kind of meshes. Although we cannot publish a detailled analysis, we can report that the more experienced engineers find false positives hindering efficient review. On the other hand, the novices accept and work along the predictions made by the machine learning model, delivering mesh qualities accepted by the remitting departments.

\section{Discussion and Future Work} \label{sec:discussion}
Our work demonstrates a data-driven approach to evaluating mesh quality for FEM. In principle, it can be integrated into a fully automated routine that incrementally re-generates low quality subsets of a mesh and re-evaluates the result, until a fixed point is reached that represents a mesh that is fit to enable accurate simulations.
%
Whilst experimental results show practical applicability, the imprecise semantics of expert evaluations hinders a more conclusive evaluation. The current practice of mesh review leaves unclear, e.g., whether an individual element that was marked for rework was also central to the decision about mesh quality, or whether it is merely included into a set of elements that requires re-meshing as a consequence of other (e.g., more distant) elements rendering the mesh undesirable.
The immediate ramifications are two-fold: For one, the resulting label noise can misguide the training of machine learning models to pay undue attention to inconspicuous elements that are labelled for rework and that, because of similar properties reflected in the feature vector, can be easily confused with elements that have passed expert evaluation. This can drive false positives. For another, the imprecise semantics limits applicability of element-level performance metrics. In particular, inconspicuous elements might rightly be predicted as passed, but their ground-truth label can include them into rework of a much larger undesirable structure. This can drive false negatives.

Given this, we have argued that our experiments return unfavourable results that may not accurately reflect the goal of identifying subareas of a mesh that need to be re-generated for better quality, whilst not distracting attention to inconspicuous subareas at the same time. Work in progress considers the development of a domain-specific cost function and performance metric for guiding the optimisation of a machine learning model that reflects this goal.
%
In order to address the problem of subjective labeling, we also contemplate to change the process of mesh reviews, including the automation of the labelling process by comparing meshes before and after correction.

Whilst a more domain-specific metric or clearer labels might improve model optimisation and interpretability of the results, we also expect potential improvements with the inclusions of additional low-level features, and alternative machine learning techniques that might better abstract to higher-level representations.
For instance, some additional low-level features could relate an element to the underlying geometry of the original design. The original geometry was not available for our experiments, but might have influenced the review decisions by expert engineers. Of particular relevance are so-called feature lines in the geometry, i.e., lines which define the structure of a part like bends from manufacturing processes or important design features.
Promising approaches that might capture dependencies across larger neighbourhoods, i.e., beyond the scope of $k$-ring neighbourhoods for small~$k$, include convolutional neural networks and graph neural networks.

Convolutional neural networks~(CNNs, \cite{lehabobe99}) aim at learning high-level representations from low-level features through kernels that slide along a feature tensor, and have revolutionalised the field of computer vision with their ability to learn to recognise structures within pixel data. Research related to our work has applied CNNs to the problem of mesh classification, i.e., the tasks of identifying the object represented by a mesh or assigning each mesh element to an object part that the element belongs to. The work presented in~\cite{kadoxi16a} organises low-level features of a mesh in a two-dimensional grid and reports impact from the application of CNNs. We have tested the application of CNNs on our feature tensor but have not found them to improve performance over FNNs. Spin images~\cite{johe99} provide an alternative to aggregating data from $k$-ring neighbourhood frontiers that, in contrast to our approach, can accumulate neighbourhood data by rotating a rectangular sheet along the surface normal through the centre of mass of an element. We have also tested the application of CNNs when spin images are used to aggregate neighbourhood statistics, but have not found them to improve performance over the feature vector presented in this paper.
%
Another, more na\"ive avenue to making CNNs applicable for element classification is to generate image data via mesh rendering, and using a CNN to solve an image segmentation problem. Image segmentation partitions the resulting image into segments, e.g., ones that might require rework and ones that pass. However, mesh rendering can result in a significant loss of information because there exists no two-dimensional projection of three-dimensional surfaces that retains all relevant properties, such as lengths, angles, and area. In fact, we have tested this approach at an earlier stage of our research with very limited success.

Recent advances in graph neural networks~(GNNs, \cite{gomosc05}), including graph convolutional networks~\cite{kiwe17}, graph attention networks~\cite{vecucarolibe18}, and gated graph neural networks~\cite{litabrze16}, have led to ground-breaking results in learning tasks that require dealing with graph data. Unlike our approach, i.e., aggregating features extracted from each element neighbourhood to allow the application of standard neural networks, a GNN can represent information from a neighbourhood with arbitrary depth. We believe that the latter is required to capture undesirable mesh structures that are beyond the reach of our current representation, e.g., stretches of quadrilaterals that are enclosed by an opening and closing triangle.
To our knowledge, classification models for graph-data are mainly studied in semi-supervised learning tasks that naturally arise in online social networks domains~\cite{aggarwal2011}. In semi-supervised learning, some of the data is labelled and the task is to find labels for unlabelled data, e.g., by propagating structural information and attributes in a graph to determine the label of unlabelled vertices from some labelled ones.
This contrasts to our application scenario, i.e., a supervised learning task, where no labels are available to the elements from unseen meshes. Hence, our work contributes a novel, real-world application scenario and benchmark domain to the study of GNNs.
Another, more indirect and speculative alternative is the hybridisation of our approach by considering only a conservative selection of elements for rework, e.g., through higher thresholds, and treating the classification of the remaining elements of a mesh as a semi-supervised task for GNNs. This is left to future work.

\section{Conclusions} \label{sec:conclusions}
We have put forward a novel, automated method for postprocessing finite element meshes, a laborious and time-consuming task that is typically performed by engineering experts to this date.
A central problem to this task is the evaluation of mesh quality, e.g., the fitness of a mesh to enable accurate simulations, because many complex and sometimes subjective, and conflicting requirements from different simulation disciplines are yet to be formalised.
We address this issue by applying machine learning to abstract from expert evaluations. Our method is entirely data-driven, i.e., it relies on the ability of machine learning techniques to learn a representation of high-level concepts that capture mesh quality from low-level features.

We have explained why standard machine learning models like tree-based models and standard neural networks cannot handle mesh input directly. To allow their application, we have proposed the element-wise extraction of a set of domain-specific low-level properties by means of the neighbourhood graph induced by a mesh. Experimental results from evaluating the quality of finite element shell meshes for the purpose of structural mechanic simulations demonstrate practical applicability with using off-the-shelf machine learning techniques, including extremely randomised trees and feedforward neural networks, albeit an objective evaluation proves tricky in this subjective problem domain.

Potential improvements include the development of a domain-specific cost function and performance metric to better guide the training of machine learning models, the inclusion of geometry-related features, and alternative machine learning methods that better abstracts to high-level concepts of mesh quality.
Future work considers potential benefits from the application of graph neural networks, a field that has recently seen considerable advances. We believe that element classification contributes a novel, challenging and practical application scenario to its research, which has previously focussed primarily on social graphs.


\paragraph*{Acknowledgements} 
This research was supported by the German Federal Ministry of Education and Research
(BMBF) via project AIAx-Machine Learning-driven Engineering (Nr. 01IS18048).
\bibliography{meshing}
\bibliographystyle{elsarticle-num}
\end{document}